\documentclass{article}
\usepackage{spconf,amsmath,graphicx}

\usepackage{enumitem}
\setlist{nosep, leftmargin=14pt}

\usepackage{mwe} 
\usepackage{multirow}
\usepackage{mathrsfs}
\usepackage{amssymb}
\usepackage{hyperref}

\newcommand{\xmark}{\tiny-}

\title{MATIS: Masked-Attention Transformers for Surgical Instrument Segmentation}
\name{Nicolás Ayobi, Alejandra Pérez-Rondón, Santiago Rodríguez, Pablo Arbeláez}
\address{Center for Research and Formation in Artificial Intelligence \\
Universidad de los Andes, Colombia
}
\begin{document}

\maketitle

\ninept
\begin{abstract}

We propose Masked-Attention Transformers for Surgical Instrument Segmentation (MATIS), a two-stage, fully transformer-based method that leverages modern pixel-wise attention mechanisms for instrument segmentation. MATIS exploits the instance-level nature of the task by employing a masked attention module that generates and classifies a set of fine instrument region proposals. Our method incorporates long-term video-level information through video transformers to improve temporal consistency and enhance mask classification. We validate our approach in the two standard public benchmarks, Endovis 2017 and Endovis 2018. Our experiments demonstrate that MATIS’ per-frame baseline outperforms previous state-of-the-art methods and that including our temporal consistency module boosts our model’s performance further. Our code can be found at \url{https://github.com/BCV-Uniandes/MATIS}.
\end{abstract}
\section{Introduction}
\label{sec:intro}


Instrument segmentation is critical for surgical scene understanding, as it enables the development of computer-assisted systems for instrument tracking \cite{chmarra2007systems,laina2017concurrent}, pose estimation \cite{du2018articulated}, and surgical phase estimation \cite{sanchez2022data}. There has been significant progress in addressing this problem, however there are two domains that still require improvement. The first is accurately identifying and segmenting surgical instruments, which can be challenging due to the high similarity between multiple instrument types and the class imbalance in available datasets. The second is incorporating temporal information in videos to enable consistent recognition among frames and improve overall understanding of the surgical procedure.

Models based on Convolutional Neural Networks (CNNs) have been common for surgical instrument segmentation. Initial efforts used Fully Convolutional Networks (FCN) for instrument parts segmentation \cite{toolnet,endovis2015}. Later, the Endoscopic Vision (Endovis) 2017 Robotic Instrument Segmentation Challenge \cite{endovis2017} and its 2018 version \cite{endovis2018} introduced the instrument sub-type segmentation task. Thus, most methods adapted FCN-based models to perform semantic segmentation of the different instruments present on each frame \cite{endovis2017,endovis2018,ternausnet,unetplus}. Some models leverage additional priors like optical and motion flow \cite{dualmf,mftapnet}, stereoscopic information \cite{stereo}, or saliency maps \cite{saliency}. More recent approaches have modified the original task by including weak supervision \cite{dualmf,sanchez}, domain adaptation \cite{onetomany}, pose estimation \cite{du2018articulated}, kinematic data \cite{sintetic,kinematic_padoy}, and image generation \cite{image_to_image,sintetic}. 

However, these previous models use a per-pixel classification strategy that categorizes each pixel into one of the instrument types. Hence, these methods often lack spatial consistency and disregard the multi-instance nature of the instrument segmentation problem. Therefore, ISINet \cite{isinet} first formulated a region classification approach for instance-based instrument segmentation using Mask-RCNN \cite{maskrcnn} followed by optical flow to improve temporal consistency. Alternatively, \cite{offsets} approaches instrument segmentation by simultaneously predicting each pixel's class and instance number.

\begin{figure}[t]
    \centering
    \includegraphics[width=\linewidth]{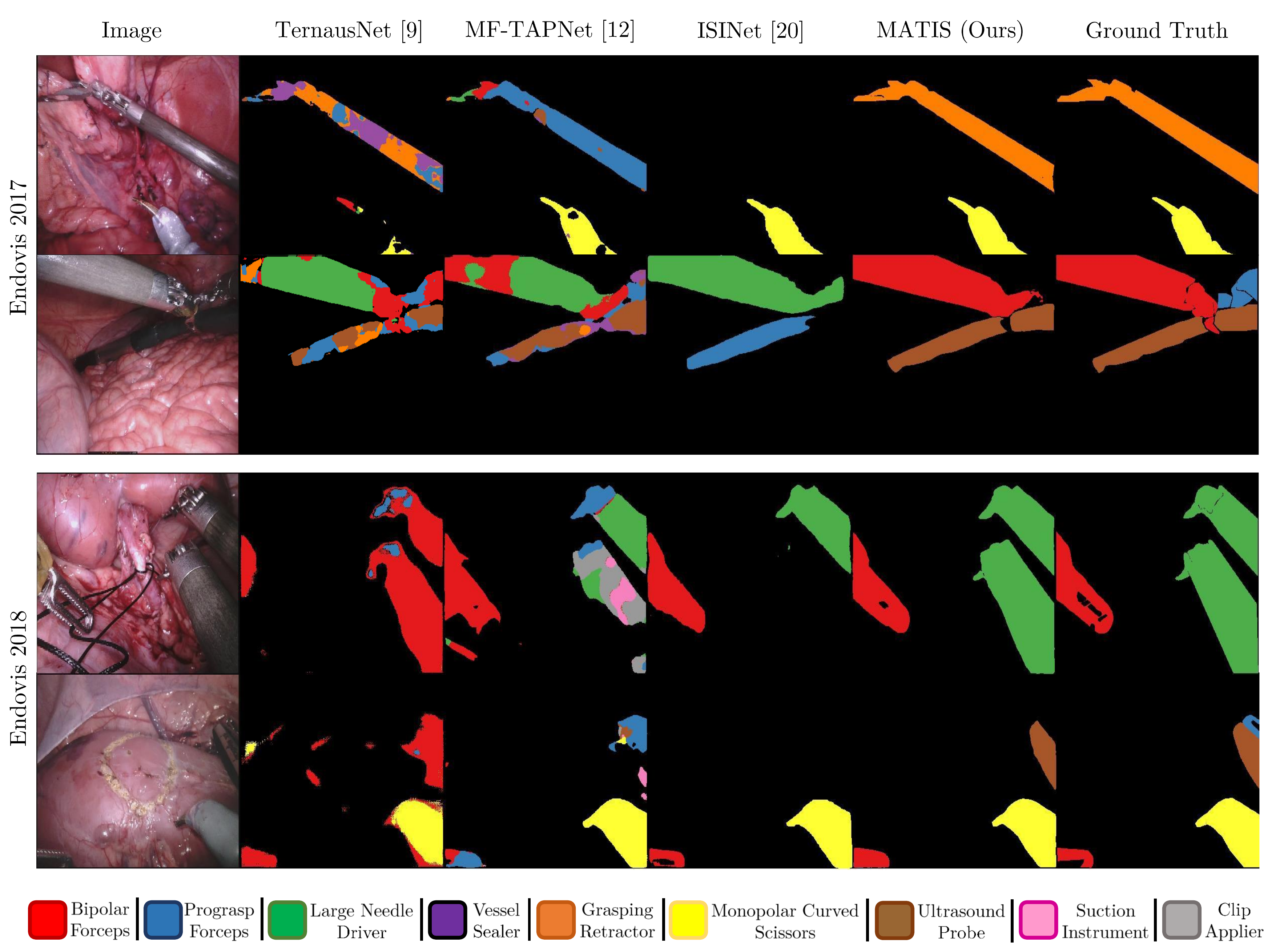}
\caption{\textbf{Qualitative comparisons of MATIS} with previous methods in the EndoVis 2017 and EndoVis 2018 datasets.}
    \label{fig:qualitative}
\end{figure}

Recently, Vision Transformers (ViTs) \cite{swin,vit} have become state-of-the-art in multiple computer vision tasks \cite{swin,detr,maskformer}. Thus, \cite{MaskSwinCnn} first incorporated transformers for instrument segmentation by combining CNNs and Swin transformers \cite{swin} as a backbone for Mask-RCNN. Furthermore, Transformer-based segmentation architectures like DETR \cite{detr} and MaskFormer \cite{maskformer} outperformed CNN-based methods by using a fixed-size set of learnable object queries to predict a set of regions. These architectures have shown that region classification as a set prediction problem can be used for every segmentation or detection task. For instance, TraSeTr \cite{trasetr} outperformed previous instrument segmentation methods using a modified MaskFormer model with a CNN backbone. Nevertheless, recently improved architectures like Deformable DETR \cite{def_detr} and Mask2Former \cite{mask2former} demonstrated the segmentation potential of deformable attention and masked attention mechanisms over standard self and cross-attention in \cite{detr,maskformer}. These architectures achieved state-of-the-art performance in all segmentation tasks and allowed faster and easier training. Still, these architectures have not been employed for instrument segmentation.

Moreover, video transformers have also achieved state-of-the-art results, again outperforming CNN-based methods \cite{videoswin,mvit}. However, they are yet to be fully leveraged for surgical instrument segmentation. Most previously proposed CNN-based methods include optical flow \cite{mftapnet,dualmf,isinet} information rather than exploiting global video-level understanding. Furthermore, TraSeTr \cite{trasetr} lacks long-term video reasoning as it implicitly tracks instruments through time by using the output embeddings of the previous frame as an additional input query for the actual frame. Conversely, STSwinCL \cite{stswincl} adapts a Swin transformer for video analysis with a contrastive learning approach for surgical scene semantic segmentation. However, this method uses a pixel-classification modality that does not differ among instances. Recently, TAPIR \cite{tapir} addressed these problems and posed a fully transformer-based model for multi-level surgical workflow analysis that incorporates Deformable DETR \cite{def_detr} as a box proposal network for instrument detection.

We propose Masked-Attention Transformers for Instrument Segmentation (MATIS), a transformer-based model that uses \textit{masked and deformable attention} to perform instrument instance segmentation and employs \textit{video transformers} to enhance the mask classification process further. MATIS utilizes Mask2Former \cite{mask2former} with a Swin Transformer backbone \cite{swin} as an instance segmentation baseline to leverage its \textit{multi-scale deformable attention pixel decoder} and \textit{masked attention mechanisms}. Furthermore, MATIS uses a TAPIR-like \cite{tapir} architecture to include temporal information and classify the regions proposed by its masked attention baseline. In summary, our key contributions are:
\begin{itemize}
    \item We propose a fully transformer-based mask classification architecture that uses masked and deformable attention for instance-based instrument segmentation.
    \item We incorporate video transformers for instrument segmentation to fully exploit temporal information.

\end{itemize}
MATIS is a novel architecture that achieves higher overall performance than all previous instrument-type segmentation methods, thus stating a new state-of-the-art for surgical instrument segmentation. Our training and validation code, including our pretrained models, can be found at \url{https://github.com/BCV-Uniandes/MATIS}.
\section{MATIS}
\label{sec:method}

\begin{table*}[h]
\centering
\resizebox{\textwidth}{!}{
\begin{tabular}{cccccccccccc}
\hline
\multirow{2}{*}{Dataset}                          & \multirow{2}{*}{Method} & \multirow{2}{*}{mIoU} & \multirow{2}{*}{IoU} & \multirow{2}{*}{mcIoU} & \multicolumn{7}{c}{Instrument Categories}             \\ \cline{6-12}
                                                  &                         &                       &                      &                        & BF    & PF    & LND   & VS/SI & GR/CA & MCS   & UP    \\ 
\hline
\hline
\multicolumn{1}{c}{\multirow{8}{*}{Endovis 2017}} & TernausNet \cite{ternausnet}   & 35.27 & 12.67 & 10.17 & 13.45 & 12.39 & 20.51 & 5.97  & 1.08  & 1.00  & 16.76 \\
\multicolumn{1}{c}{}                              & MF-TAPNet \cite{mftapnet}   & 37.25 & 13.49 & 10.77 & 16.39 & 14.11 & 19.01 & 8.11  & 0.31  & 4.09  & 13.40 \\
\multicolumn{1}{c}{}                              & Dual-MF \cite{dualmf,trasetr}     & 45.80 $\pm 21.45$ &       & 26.40 & 34.40 & 21.50 & 64.30 & 24.10 & 0.8   & 17.90 & 21.80 \\
\multicolumn{1}{c}{}                              & ISINet  \cite{isinet}     & 55.62 & 52.20 & 28.96 & 38.70 & 38.50 & 50.09 & 27.43 & 2.1   & 28.72 & 12.56 \\
\multicolumn{1}{c}{}                              & TraSeTr \cite{trasetr}     & 60.40 $\pm 4.80$ &       & 32.56 & 45.20 & \textbf{56.70} & \textbf{55.8}  & \textbf{38.90} & 11.40 & \textbf{31.30} & 18.20 \\
\multicolumn{1}{c}{}                              & Kurmann et. al. \cite{offsets}       & 65.70 $\pm 0.07$ &       &       &       &       &       &       &       &       &       \\

\multicolumn{1}{c}{}                              & \textbf{MATIS Frame (Ours)} & 68.79 $\pm 2.98$ & 62.74 & 37.30 & 66.18 & 50.99 & 52.23 & 32.84 & 15.71 & 19.27 & 23.90 \\

\multicolumn{1}{c}{} &
\textbf{MATIS Full (Ours)} & \textbf{71.36 $\pm 3.46$} & \textbf{66.28} & \textbf{41.09} & \textbf{68.37} & 53.26 & 53.55 & 31.89 & \textbf{27.34} & 21.34 & \textbf{26.53} \\

\hline
\multirow{6}{*}{Endovis 2018}                     & TernausNet \cite{ternausnet}  & 46.22 & 39.87 & 14.19 & 44.20 & 4.67  & 0.00  & 0.00  & 0.00  & 50.44 & 0.00  \\
                                                  & MF-TAPNet  \cite{mftapnet}  & 67.87 & 39.14 & 24.68 & 69.23 & 6.10  & 11.68 & 14.00 & 0.91  & 70.24 & 0.57  \\
                                                  & ISINet  \cite{isinet}     & 73.03 & 70.94 & 40.21 & 73.83 & 48.61 & 30.98 & 37.68 & 0.00  & 88.16 & 2.16  \\
                                                  & TraSeTR  \cite{trasetr}    & 76.20 &       & 47.71 & 76.30 & \textbf{53.30} & 46.50 & 40.60 & \textbf{13.90} & 86.20 & 17.15 \\
                                                  & \textbf{MATIS Frame (Ours)} & 82.37 & 77.01 & 48.65 & 83.35 & 38.82 & 40.19 & 64.49 & 4.32  & \textbf{93.18} & 16.17\\
                                                  & \textbf{MATIS Full (Ours)} & \textbf{84.26} & \textbf{79.12} & \textbf{54.04} & \textbf{83.52} & 41.90 & \textbf{66.18} & \textbf{70.57} & 0.00  & 92.96 & \textbf{23.13}\\ 
                                                  
                                \hline

\end{tabular}}
\caption{\textbf{Comparative results of MATIS in Endovis 2017 and Endovis 2018}. Instrument categories are Bipolar Forceps (BF), Prograsp Forceps (PF), Large Needle Driver (LND), Vessel Sealer (VS), Suction Instrument (SI), Grasping Retractor (GR), Clip Applier (CA), Monopolar Curved Scissors (MCS) and Ultrasound Probe (UP). \textit{MATIS Frame} stands for the independent masked attention baseline performance and \textit{MATIS Full} for the entire model, including the temporal consistency module. Both models significantly outperform all previous methods in overall metrics. The best results are shown in bold.}
\label{tab:results}
\end{table*}

\begin{figure}[t]
    \centering
    \includegraphics[width=\linewidth]{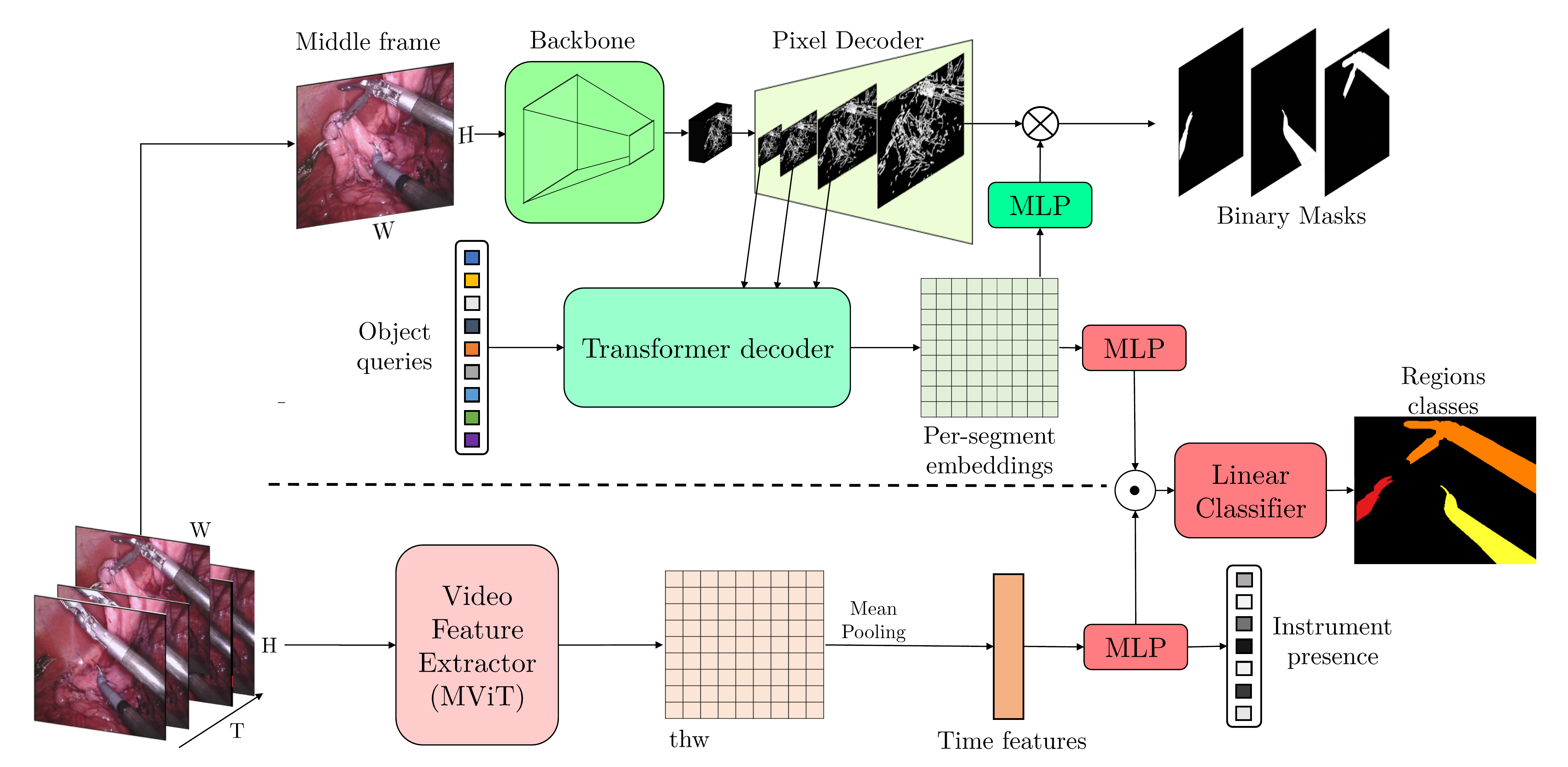}
\caption{\textbf{MATIS} first leverages Mask2Former's \cite{mask2former} meta-architecture (top) to compute a set of region proposals and their corresponding segment embeddings. MATIS' temporal consistency module (bottom) computes a sequence of spatio-temporal features that are pooled through time and linearly transformed with an MLP. The result is concatenated ($\bigodot$) with a linear transformation of the per-segment embeddings. Finally, a linear classifier predicts the final class for each region.}
    \label{fig:matis}
\end{figure}

We propose a two-stage mask classification approach for surgical instrument segmentation. Our architecture works hierarchically by predicting a set of region candidates for each instance of an instrument in each frame. These region proposals are then classified into surgical instrument types. Our approach consists of a masked attention transformer baseline trained end-to-end to create and classify region proposals. Additionally, we use an independent temporal consistency module that leverages video-level information to improve the classification of the masks generated by the baseline. Figure \ref{fig:matis} portrays our overall method.



\textbf{Masked Attention Baseline:} We use Mask2Former \cite{mask2former} as our \textit{baseline}. Mask2Former follows DETR's \cite{detr} and MaskFormer's \cite{maskformer} set prediction formulation, which aims to predict a fixed-size set $z$ of $\mathcal{N}$ class probability-binary mask pairs by using $\mathcal{N}$ input learnable queries. We refer the reader to \cite{mask2former} for a detailed explanation of Mask2Former's architecture. We modify Mask2Former to the specific task of instrument type segmentation by adapting the segmentation module to the number of classes available in our tested datasets. For inference, we use the information on the number of possible instances per instrument as priors to select the top-k scoring regions per class. In the same way, we design class-specific score thresholds to counter the high inter-class score variability caused by the class frequency miss-balance. The following section explains how we use the video analysis methodology of \cite{tapir} to improve our segment classification. \\



\noindent{\textbf{Temporal Consistency: } Following TAPIR \cite{tapir}, our \textit{temporal consistency module} uses a Multi-Scale Vision Transformer (MViT) \cite{mvit} as a backbone for video analysis. TAPIR uses a time window centered on a keyframe to compute global spatio-temporal features that encode the complex temporal context of the middle frame. Additionally, we adapt the original box classification head into a region classification head by using Mask2Former's per-segment embeddings, which capture localized region features. As in TAPIR, we pool the temporal features through time; however, we also apply a multilayer perceptron (MLP) to the pooled time features before concatenating them, and then we linearly classify each region. Finally, we include additional supervision by using another MLP on the pooled time features to predict the presence of each instrument in the middle frame. We supervise this multi-target recognition task with a binary cross-entropy loss added to the original mask classification loss.} \\

\noindent{\textbf{\textit{Implementation Details:}} For the masked attention baseline, we use Mask2Former's official implementation \cite{mask2former} pretrained in MS-COCO \cite{coco} instance segmentation. We keep $\mathcal{N}=100$ to leverage Mask2Former's object queries pretraining fully. We train Mask2Former for 100 epochs on 4 NVIDIA Quadro RTX 8000 GPUs with a batch size of 24 and an ADAMW optimizer. We use a Swin Small (SwinS) \cite{swin} backbone as it provides the best trade-off between parameters and performance. For segmentation inference, we use $k=2$ for instruments with multiple instances on a frame and $k=1$ for instruments that always have one instance.}

For the temporal consistency module, we use TAPIR's official implementation \cite{tapir} pretrained in Kinetics 400 \cite{kinetics}. We train the model for 20 epochs in a single NVIDIA Quadro RTX 8000 GPU with a window size of 8 frames, a stride of 1 frame, a batch size of 12, and an ADAMW optimizer. We use the same learning rate adjustment policy as in \cite{tapir} and the same rescaling and augmentation strategies. We only use the regions selected with the masked attention baseline inference strategy for training and validation. 


\section{Experiments and Results}
\label{sec:pagestyle}

\subsection{Datasets and Evaluation Metrics}

We train and evaluate our method in two public experimental frameworks:  the Endovis 2017 \cite{endovis2017}, and Endovis 2018 \cite{endovis2018} datasets. For a fair comparison with previous methods, we follow the same standard benchmarks established by \cite{isinet} and \cite{ternausnet}. For Endovis 2017, we use the 4-fold cross-validation presented in \cite{ternausnet}. For Endovis 2018, we use additional instance annotations given by \cite{isinet} and their predefined training and validation splits. For evaluation, we adopt the three common segmentation metrics from \cite{isinet}: Mean Intersection over Union (mIoU), Intersection over Union (IoU), and Mean Class Intersection over Union (mcIoU). Finally, we report the standard deviation error among the folds of Endovis 2017.


\subsection{Experimental validation}
\label{sec:validation}


First, we validate the individual performances of MATIS's baseline (without the temporal consistency module) against previous state-of-the-art methods. We call this experiment \textit{MATIS Frame}, and the comparative results are shown in Table \ref{tab:results}. \textit{MATIS Frame} outperforms all previous methods in all three overall segmentation metrics. We also validate our baseline in Endovis 2018 without $seq2$ for a fair comparison with \cite{sanchez}, achieving a 4.6\% IoU improvement. Our results corroborate the superiority of transformer-based and mask-classification approaches over CNN-based and pixel-classification methods. The segmentation improvement over TraSeTr demonstrates that multi-scale deformable attention and masked attention provide a more localized and flexible understanding of instruments' visual characteristics, leading to better segmentations. Regarding the class-wise metrics, MATIS strongly outperforms TraSeTr in categories with the highest amount of training annotations, even achieving over a 20\% difference in some classes. Nevertheless, the higher performance of TraSeTr in the remaining classes is due to their implicit instrument tracking during segmentation computation. 


 

\begin{table}[]
\centering
\resizebox{\linewidth}{!}{
\begin{tabular}{cccc}
\hline
Dataset                       & Upper Bound          & mIoU                      & mcIoU                     \\ \hline
\multirow{2}{*}{Endovis 2017} & \textit{Inferred Upper Bound} & 83.44                     & 81.86                     \\
                              & \textit{Total Upper Bound}    & \multicolumn{1}{c}{90.75} & \multicolumn{1}{c}{90.44} \\ \hline
\multirow{2}{*}{Endovis 2018} &\textit{ Inferred Upper Bound} & 88.84                     & 80.68                     \\
                              & \textit{Total Upper Bound }   & 91.20                     & 88.67                     \\ \hline
\end{tabular}}
\caption{\textbf{Upper bound} results for matching annotations with predictions to estimate segmentation metrics for a perfect region proposal classification. \textbf{Note:} \textit{mIoU equals IoU for a perfect mask classification.} }
\label{tab:bounds}
\end{table}

Additionally, we calculate two segmentation upper bounds to evaluate our segments' quality and quantify our mask classification errors. We calculate the \textit{Inferred Upper Bound} by finding the best IoU match within the set of inferred regions for each annotated instance. We discard the remaining inferred unmatched masks. Moreover, we calculate the \textit{Total Upper Bound} by doing the same procedure but using all the 100 predicted regions without our inference procedure. The results are shown in Table \ref{tab:bounds}. Our high upper-bound values highlight our method's capacity to produce near-perfect instrument segmentation masks and confirm that the primary source of error is region classification rather than mask generation. We can observe this behavior in our qualitative results in Figure \ref{fig:qualitative}.


Furthermore, we evaluate the effect of including our temporal consistency module (\textit{MATIS Full}) and compare it against \textit{MATIS Frame} and previous models. We also exhibit the results in Table \ref{tab:results}. Our temporal consistency module significantly boosts all the overall evaluation metrics in Endovis 2018 and all folds of Endovis 2017. Thus, including video processing combined with our baseline's localized segment embeddings corrects multiple miss-classifications caused mainly by the low interclass variability. These enhancements can be primarily seen in Endovis 2018's Large Needle Driver category, whose instances are constantly classified as Prograsp Forceps by our baseline. Still, some classes in Endovis 2018 have a performance decrease due to classification noise introduced by long-term video reasoning. Also, we observe a considerable increase in the standard deviation in Endovis 2017 caused by a variable performance increase on each fold.

Last, Figure \ref{fig:qualitative} visualizes segmentation examples. These qualitative results show our masks' superior pixel accuracy and spatial consistency against previous publicly available methods. In the same way, MATIS exhibits a higher instrument identification capacity. Nonetheless, we observe that our method struggles to segment overlapped or discontinuous instances and that it still presents some mask miss-classifications.



\subsection{Ablation Experiments}

\begin{table}[t]
\centering
\resizebox{\linewidth}{!}{
\begin{tabular}{cccc}
\hline
Inference method                                & mIoU           & IoU            & mcIoU          \\ \hline
All masks                                    & 80.71          & 52.39          & 35.22          \\
NMS                                    & 81.21          & 56.06          & 35.93          \\
0.5 threshold                                   & 81.06          & \textbf{77.15} & 45.33          \\
Top 4 instances                                        & 82.02          & 64.66          & 42.03          \\
Per-class thresholds                             & 82.29          & 76.60          & 47.92          \\
Top k per-class                                 & 82.26          & 53.20          & 36.61          \\
\textbf{Top k per-class  + Per-class thresholds} & \textbf{82.37} & 77.01          & \textbf{48.65} \\
\hline
\end{tabular}}
\caption{\textbf{Ablation results in MATIS Frame inference in Endovis 2018.} We considered multiple filtering and selection strategies. Combining class-specific inference methods improves MATIS' performance. The best results are shown in bold.}
\label{tab:inference}
\end{table}

\textbf{Baseline Inference:} Table \ref{tab:inference} shows the results of the multiple procedures considered for MATIS' baseline inference in Endovis 2018. We validate that our class-specific region selection method is optimal for instrument segmentation inference. We test using all the 100 predicted masks (\textit{All masks}), filtering with 0.5 confidence score threshold (\textit{0.5 threshold}), selecting the top 4 scoring masks (\textit{Top 4}), the no-max suppression (\textit{NMS}) strategy in \cite{trasetr}, and multiple combinations of the previous processes. However, we observe that classes with fewer training instances produce masks with low confidence scores. Hence, per-class selection and per-class thresholds perform better than general filtering. 

\noindent{\textbf{Temporal consistency module design:} On the one hand, we verify the effect of our additional multilayer perceptron for the pooled time features (\textit{Time MLP}), and the use of multi-label classification for instrument presence supervision (\textit{Presence Supervision}). The results are presented in Table \ref{tab:tapir}. Our results demonstrate that the time consistency module achieves higher performance than the sole baseline without modifications. However, the inclusion of the \textit{Time MLP} makes time features more compatible with segment embeddings for better joint classification. In the same way, additional recognition supervision allows time features to capture richer information about instrument types. Thus, both modifications improve our metrics further, and the best performance is achieved by simultaneously applying them. On the other hand, we validate our input size hyperparameters by testing multiple input window sizes and stride values. We observe that performance decreases with smaller window sizes, hence proving the advantages of long-term information. Similarly, increasing the stride value diminishes our performance due to the low sampling rate in our datasets. Finally, larger window sizes slightly diminish performance (less than 1\%) as they introduce too much noise from far frames.}

\begin{table}[t]
\centering
\resizebox{\linewidth}{!}{
\begin{tabular}{ccccc}
\hline
\multicolumn{2}{c}{Attribute}     & \multirow{2}{*}{mIoU} & \multirow{2}{*}{IoU} & \multirow{2}{*}{mcIoU} \\ \cline{1-2}
Time MLP & Presence Supervision &                       &                      &                        \\ \hline
\xmark        &  \xmark                      &  \multicolumn{1}{c}{83.24}  & \multicolumn{1}{c}{78.10} & \multicolumn{1}{c}{50.31}   \\
\checkmark        & \xmark                       &     83.50                  & 78.31            &     51.21                  \\
\xmark        &  \checkmark                      &       83.63                &         78.23             &        50.96                \\
\checkmark        & \checkmark                       &    \textbf{84.26}             &         \textbf{79.12}             &       \textbf{54.02}   \\ \hline             
\end{tabular}}
\caption{\textbf{Ablation results for TAPIR's modifications in Endovis 2018.} Time MLP and the Presence Supervision attributes refer to linearly projecting the pooled time features and using multi-label classification for instrument presence supervision, respectively. Both attributes improve MATIS' performance. The best results are shown in bold. \checkmark denotes an attribute's presence, and $\xmark$ denotes absence.}
\end{table}
\label{tab:tapir}

\section{Conclusions}
\label{sec:conclusions}

We propose a novel transformer-based architecture for surgical instrument segmentation that exploits global temporal information. Our method, MATIS, uses a masked attention baseline that leverages multi-scale deformable attention, masked attention, and our proposed class-specific inference method to predict a set of binary instrument instance masks. We employ a temporal consistency module that combines long-term video features with localized segment embeddings to improve region classification. MATIS' sole baseline outperforms previous methods in the Endovis 2017 and 2018 datasets. Our experiments show that our region proposals have a very high pixel accuracy but still have room for improvement in classification. Finally, our Time Consistency Module boosts our segmentation metrics by improving classification consistency through time. Hence, MATIS establishes a new state-of-the-art performance for surgical instrument segmentation that enables a finer and more consistent understanding of robot-assisted surgical scenes. \\

\noindent{\textbf{Compliance with ethical standards:} Our research was conducted using publicly available data from the MICCAI's Endoscopic Vision Challenge \cite{endovis2017,endovis2018}. Since these are standard databases widely used in the field, the data collection has been revised to meet all ethical requirements, and its use does not require additional ethical approvals.} 


\noindent{\textbf{Acknowledgments:} Nicolás Ayobi, Alejandra Pérez-Rondon, and Santiago Rodríguez acknowledge the support of the 2022 and 2023 UniAndes-DeepMind Scholarships.}

\bibliographystyle{IEEEbib}

\bibliography{refs}

\end{document}